\documentclass[sigconf]{acmart}
\renewcommand\footnotetextcopyrightpermission[1]{} 

\AtBeginDocument{%
  \providecommand\BibTeX{{%
    \normalfont B\kern-0.5em{\scshape i\kern-0.25em b}\kern-0.8em\TeX}}}

\usepackage{subfigure}
\usepackage{balance}

\setcopyright{none}
\begin{document}
\fancyhead{}
\pagenumbering{gobble}

\title{Cobbler Stick With Your Reads}
\subtitle{People's Perceptions of Gendered Robots Performing Gender Stereotypical Tasks}

\author{Sven Neuteboom}
\affiliation{%
  \institution{Utrecht University}
  \city{Utrecht}
  \country{The Netherlands}}
\email{s.y.neuteboom@student.uu.nl}

\author{Maartje M.A. de Graaf}
\affiliation{%
  \institution{Utrecht University}
  \city{Utrecht}
  \country{The Netherlands}}
\email{m.m.a.degraaf@uu.nl}


\begin{abstract}

Previous research found that robots should best be designed to fit their given task, whilst others identified gender effects in people’s evaluations of robots. This study combines this knowledge to investigate stereotyping effects of robot genderedness and assigned tasks in an online experiment (\textit{n} = 89) manipulating robot gender (male vs. female) and task type (analytical vs. social) in a between subject’s design in terms of trust, social perception, and humanness. People deem robots more competent and have higher trust in their capacity when they perform analytical tasks compared to social tasks, independent of the robot's gender. Furthermore, we observed a trend in the data indicating that people seem to dehumanize female robots (regardless of task performed) to animals lacking higher-level mental processes, and additionally that people seem to dehumanize robots to emotionless objects only when gendered robots perform tasks contradicting the stereotypes of their gender.

\end{abstract}






\maketitle

\section{Introduction}

 Robots will soon be embracing a myriad of tasks in our everyday lives, in hospitals, schools, the office, and our homes. Previous research in human-robot interaction (HRI) have considered different angles investigating people’s evaluations of robots’ suitability to perform a given task. Some studies analyzed people’s general social acceptance of robots in several potential future jobs \cite{graaf2016anticipating,enz2011social}. These studies conclude that people are willing to accept robots as entertainment or personal service and in applications for hazardous environments but are likely to reject robot applications that require sophisticated social emotional interactions. Other scholars specifically directed their research at a fit between task and appearance indicating that robot job suitability is both affected by expectations about a robot’s capacities needed for that job \cite{lee2011effects} as well as a clear match between a robot’s appearance and its intended functional purpose \cite{graaf2015evaluation,goetz2003matching}. Such results can be explained by a body of research in human psychology indicating that first impression formations based on appearance cues are ample triggers for social categorization \cite{bargh1999cognitive} but also prompt subsequent stereotyping processes \cite{mccauley1995stereotype, tajfel1971social}. HRI research shows that people also easily infer gender from a robot’s appearance \cite{jung2016feminizing} which trigger gender stereotypical beliefs about such gendered robots \cite{eyssel2012s,tay2014stereotypes}. The current study expands current knowledge on appearance-task fit and gender inferences by investigating stereotyping effects of robot genderedness and assigned tasks in an online experiment.


\subsection{Social Categorization and Stereotypes}
Social categorization is a cognitive process to make sense of the social world by simplifying and systematizing perceptive information \cite{bargh1999cognitive}. Such categorization serves as a beneficial heuristic when meeting strangers as we infer interpersonal characteristics based on the social group that stranger belongs to \cite{mccauley1995stereotype}. However, categorizing others to social groups rather than treating them as unique individuals has various negative consequences. Social categorization prompts our tendency to hold distort perceptions and exaggerate the differences between people from distinct social groups while perceiving intensified similarities of members within those groups \cite{tajfel1971social}. Consequently, it becomes easy to apply our distort perceptions to individual members of social groups without having to consider whether the characteristics pertain to that individual. This process is called stereotyping, which are over-generalized assessments of an individual based on the group to which that individual belongs \cite{heilman2012gender}. Gender stereotypes are automatically activated immediately following categorization of a target as a member of that group \cite{devine1989automatic}.

A long history of research on gender stereotyping shows that people tend to associate different traits to men and women. Stereotypical male traits focus on competence and agency \cite{spence1979masculinity} and denote achievement orientation (e.g., competent, ambitious), inclination to take charge (assertive, dominant), autonomy (e.g., independent, decisive) and rationality (e.g., analytical, objective) \cite{heilman2012gender}. Stereotypical female traits, on the other hand, aim at warmth and expressiveness \cite{spence1979masculinity} and denote concern for others (e.g., kind, caring), affiliative tendencies (e.g., friendly, collaborative), deference (e.g., obedient, respectful) and emotional sensitivity (e.g., intuitive, understanding) \cite{heilman2012gender}. Bem \cite{bem1974measurement} mapped this distinction between stereotypical male and female traits; a division which shows a strong overlap with the Stereotype Content Model \cite{cuddy2008warmth} dimensions of warmth and competence. Subsequent research shows that people generally deem competence more desirable for males, and warmth for females \cite{broverman1972sex}. Following the Computers Are Social Actors paradigm \cite{nass1994computers}, gender stereotypes have also been reported in HRI research.

\subsection{Gender Stereotypes in HRI Research}
People project social categories and social behaviors onto robots based on their traits and characteristics, including gender cues from physical appearance \cite{eyssel2012s} as well as face and voice \cite{powers2006advisor}. Robots were originally meant solely to perform instrumental tasks \cite{wang2018mind}. While their purpose regarding their task type is changing, this classical image of robots in dirty, dangerous and dull tasks is persisting in people's minds \cite{loughnan2007animals}. Nonetheless, human-robot collaborations increasingly become everyday practice \cite{kaniarasu2014effects}. Research on human-robot collaborations stress the importance of trust for a successful introduction of robots to the workforce \cite{hancock2011meta} as trust determines people’s willingness to work with the robot in future endeavors \cite{you2018human}. Trust as both a theoretical concept and an empirical measure has been frequently debated in human-robot interaction research, but there seems to be consensus for a dichotomous dimension of trust. On the one hand, people may trust a robot based on its capacity or reliability, and on the other hand based on its integrity or morality \cite{gaudiello2016trust,ullman2019measuring}). These trust dimensions can be linked back to gender stereotypes. Female stereotypical characteristics, such as “loyal” and “compassionate” \cite{bem1974measurement}, better fit the items measured by moral trust, such as sincerity, genuineness and ethicality \cite{patterson1984empathy}. Male stereotypical characteristics, such as “ambitious” and “self-reliant” \cite{bem1974measurement}, better fit the items measured by capacity trust, such as “competent” and “skilled” \cite{hawley2007social}. We therefore hypothesize that \textit{people have higher trust in robots that perform tasks "fitting to their gender"} (H1).

Other HRI studies specifically focus on the interaction effects between robot gender and occupational domain. People will not only more easily accept robots that are in line with existing gender-stereotypes regarding gender-task fit \cite{tay2014stereotypes}, but also will perform less well (i.e., higher error rate) during a collaborative task when our social schema for gender-task fit are being violated \cite{kuchenbrandt2014keep}. Such findings can be related to similar results in psychology indicating that occupational roles are reliably stereotyped along the social perception dimensions of warmth and competence \cite{he2019stereotypes}, which in turn have been linked to gender-stereotypical traits \cite{bem1974measurement}. Given the strong underlying social schema regarding the appearance-task fit in HRI research \cite{lee2011effects}, we expect the effects of the gendered embodiment to be dominating and therefore hypothesize that \textit{robot gender affects people's social perception of a robot, independent of performed task} (H2).

There is growing body of research investigating human misconduct with robots in terms of discrimination (e.g., \cite{bartneck2018robots}) and mistreatment (e.g., \cite{keijsers2018mindless}). Another form of human misconduct happens specifically to robotic agents with female body characteristics, namely objectification. A study investigating the discourse between learners and a female-gendered virtual tutor shows a frequent objectification of that virtual agent while placing it in a subordinate and inferior role \cite{veletsianos2008sex}. Another study explored the online commentary on videos of highly humanlike robots revealing a pervasive and unabashed objectification of female-gendered robots \cite{strait2017public}. The sexual objectification of the female body has a long tradition in psychological research \cite{loughnan2007animals} indicating that both men and women perceive sexualized women as lacking mental capacity and moral status \cite{kellie2019drives}. Combining this literature on female objectification with the gender-stereotypical expectations regarding occupational suitability of gendered robots \cite{tay2014stereotypes}, we hypothesize that \textit{people's perceptions of a robot's humanness is a function of both robot gender and performed task} (H3).

\section{Method}
To investigate stereotyping effects of robot genderedness and assigned tasks on social perception, trust, and humanness, we conducted an online experiment (\textit{n} = 89) manipulating robot gender (male vs. female) and task type (analytical vs. social) in a between subject’s design.

\begin{figure}[!htbp]
    \centering
    \captionsetup{justification=centering}
    \subfigure[Male robot]{\includegraphics[scale=0.035]{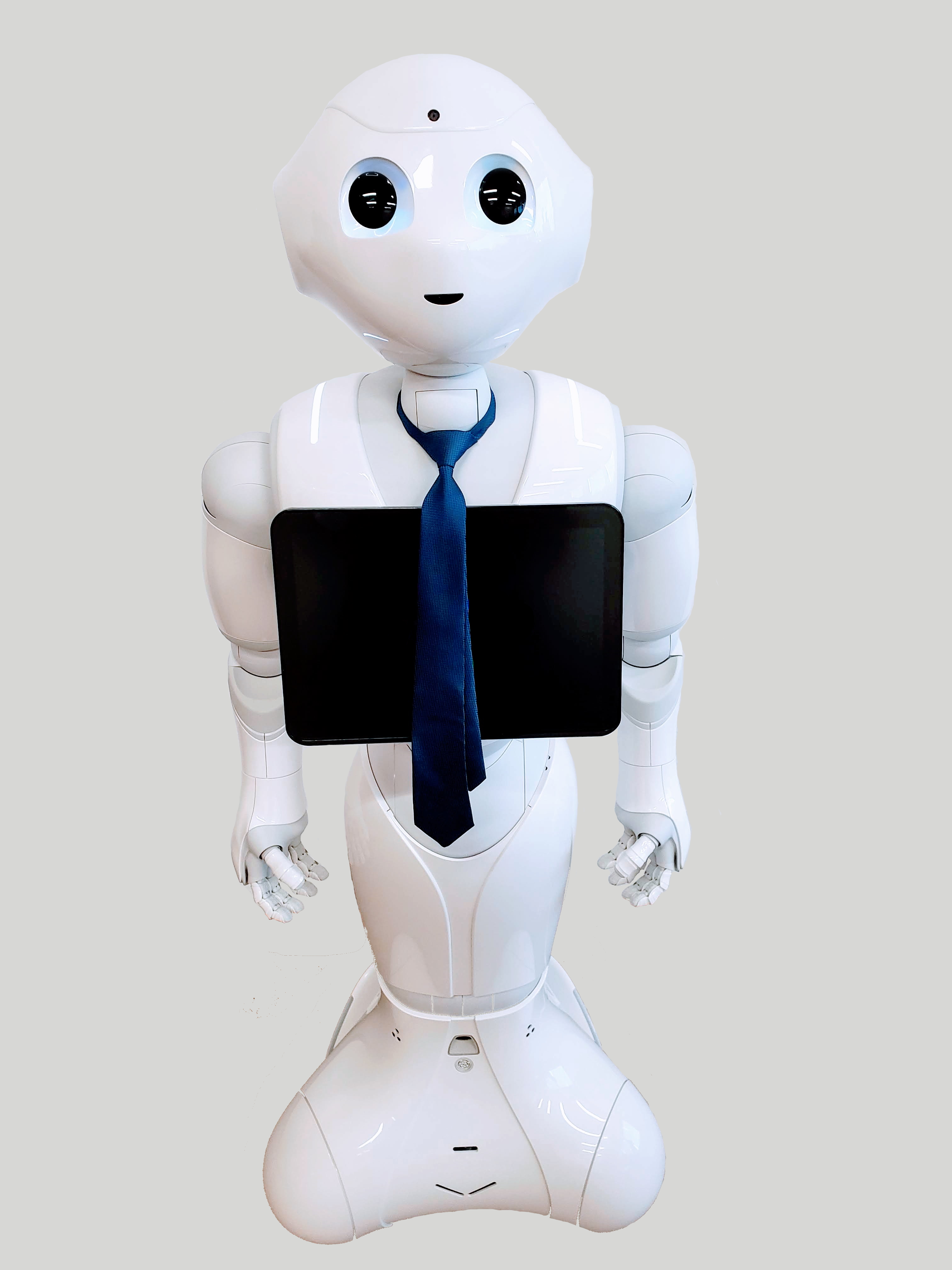}}\quad
    \subfigure[Female robot]{\includegraphics[scale=0.035]{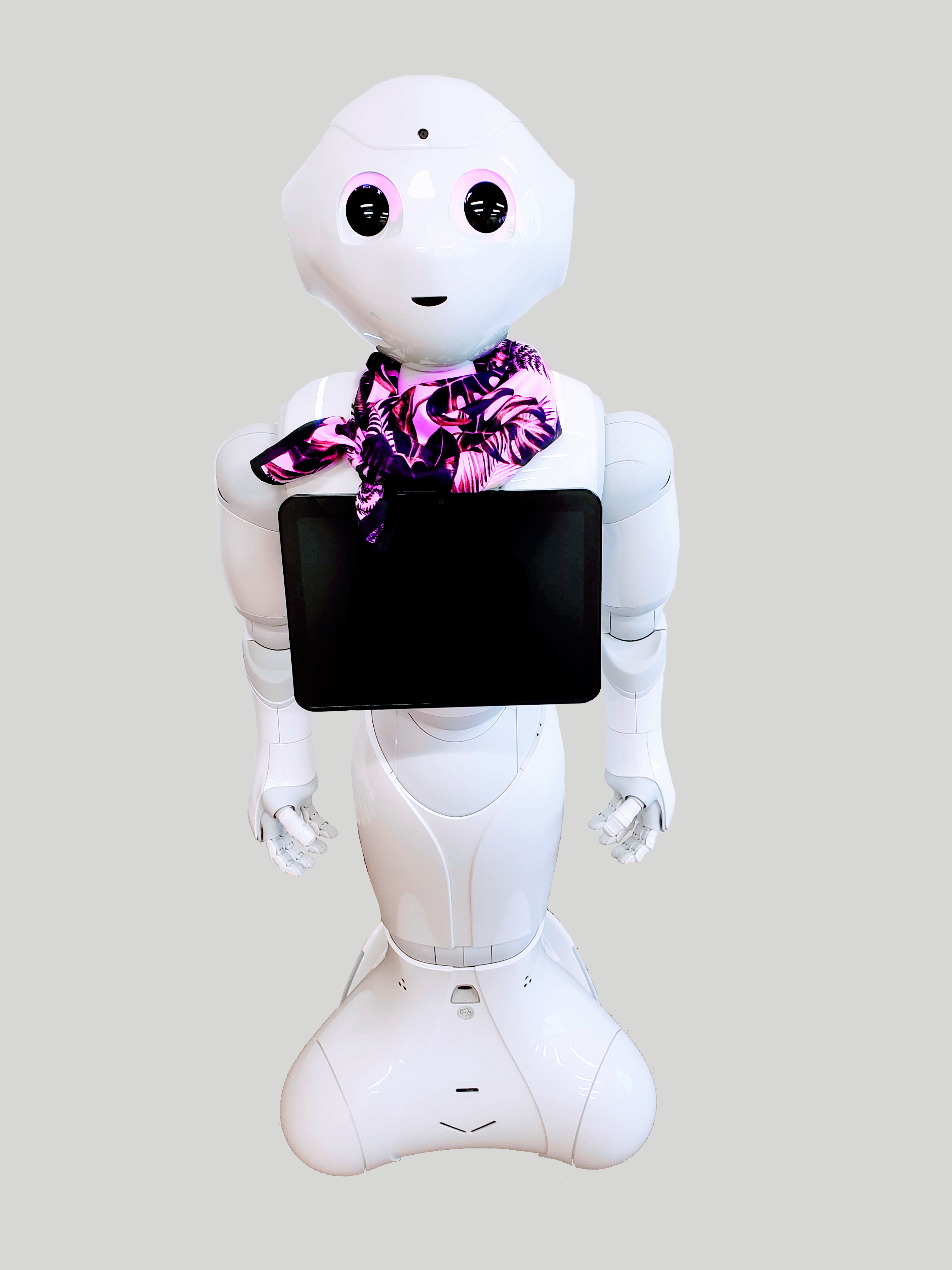}}
    \caption{Robot gender manipulation}
    \label{fig:robotgender}
\end{figure}

\begin{table*}[!htbp]
\centering
\begin{tabular}{c c c c c c}
\toprule
\multicolumn{2}{c}{ } & \multicolumn{4}{c}{\bfseries Condition}\\
\multicolumn{2}{c}{ } & \multicolumn{2}{c}{Male robot} & \multicolumn{2}{c}{Female robot}\\
\multicolumn{2}{l}{ } & Analytical task & Social task & Analytical task & Social task\\
\multicolumn{2}{l}{\bfseries Dependent variables} & \textit{Means (SD)} & \textit{Means (SD)} & \textit{Means (SD)} & \textit{Means (SD)}\\
\midrule
\multicolumn{2}{l}{Trust} & & & & \\
\multicolumn{2}{l}{\textit{Capacity trust}} & 5.09 (1.00) & 4.70 (0.94) & 4.90 (0.79) & 4.46 (0.83)\\
\multicolumn{2}{l}{\textit{Moral trust}} & 4.31 (0.92) & 4.13 (1.02) & 4.06 (0.96) & 3.93 (1.15)\\
\multicolumn{2}{l}{Social perception} & & & & \\
\multicolumn{2}{l}{\textit{Warmth}} & 4.18 (1.03) &	4.29 (1.24) & 3.88 (1.13) & 4.00 (1.42)\\
\multicolumn{2}{l}{\textit{Competence}} & 4.41 (0.88) & 4.04 (1.11) & 4.46 (0.81) & 3.63 (1.34)\\
\multicolumn{2}{l}{Humanness} & & & & \\
\multicolumn{2}{l}{\textit{Human uniqueness}} & 4.21 (1.21) & 4.17 (1.16) & 3.87 (0.89) & 3.71 (1.19)\\
\multicolumn{2}{l}{\textit{Human nature}} & 3.32 (0.93) & 3.17 (1.08) & 2.99 (1.06) & 3.68 (0.83)\\
\bottomrule
\end{tabular}
\caption{Means and Standard Deviations of Dependent Variables in Each Condition}
\label{tab1}
\end{table*}

\subsection{Stimuli}
For this experiment, we manipulated the robot gender as well as the task type to create four different vignettes. Robot gender was manipulated by modifying a picture of the Pepper robot by either giving it a blue tie for the male or a pink scarf for the female robot (see Figure \ref{fig:robotgender}), which are subtle but powerful gender cues \cite{jung2016feminizing}. Additionally, we referred to the robot as either \textit{Alexander} in the male or \textit{Alexandra} in the female task description respectively. Task type was manipulated by altering some words in a text description to indicate either an analytical or social task which were kept at similar length (i.e., 69 and 67 words respectively). The analytical task described the robot studying large datasets with medical data to provide an overview of treatment plans for hospital patients to support healthcare professionals in making solid decisions of patient treatment. The social task described the robot utilizing large datasets with verbal and non-verbal behaviors to provide emotional support to hospital patients facilitating healthcare professionals in monitoring patient well-being. The mixture of stimuli (robot gender X task type) resulted in four different vignettes.

We pretested these stimuli (\textit{n} = 12). The female robot (\textit{M} = 7.67) was perceived as more female than the male robot (\textit{M} = 5.56) measured on a 9-point Likert scale from mostly male to mostly female (\textit{p} = .012). The analytical task (\textit{M} = 8.22) was perceived as more analytical (\textit{p} = .032) than the social task (\textit{M} = 6.78), and the social task (\textit{M} = 6.67) was perceived as more social (\textit{p} < .001) than the analytical task (\textit{M} = 2.22) measured on two separate 9-point Likert scales from not at all [analytical / social] to very [analytical / social].

\subsection{Procedure}
After giving consent, participants were introduced to the survey topic by addressing the ageing society and that robots could aid the growing demand for optimization in healthcare. They were then assigned to one of the four vignettes that presented a picture of the robot (male or female) together with the task description (analytical or social), after which they responded to several statements regarding their trust, dehumanization, and social perception of the robot. The questionnaire ended with some demographic items and thanking the participant for their contribution.

\subsection{Measures}
Participants’ social perception of the robot was measured with the scale developed by Cuddy et al. \cite{cuddy2008warmth} consisting of the dimensions of warmth ($\alpha$ = .69) and competence ($\alpha$ = .67). To measure the participants’ trust in the robot, we administered the Multi-Dimensional-Measure of Trust (MDMT) scale developed by Ullman \& Malle \cite{ullman2019measuring} consisting of the dimensions of capacity trust ($\alpha$ = .77) and moral trust ($\alpha$ = .78). Perceptions of the robot’s humanness were collected using the scale by Haslam et al. \cite{haslam2006dehumanization} consisting of the dimensions of human uniqueness ($\alpha$ = .68 after removing the item of logical) and human nature ($\alpha$ = .67 after removing the item of individual). All measures have been presented on 7-point Likert scales. Table \ref{tab1} presents the means and standard deviations of all these dependent variables for each of the conditions independently.

\subsection{Participants}
A total of 95 participants were recruited via various social media. We deleted 6 responses from further analyses due to a completion rate below 75\%. Thus, we analyzed the data from 89 participants (52\% males), with age ranging from 18 to 79 (\textit{M} = 29.1, \textit{SD} = 14.4). They indicated having an average knowledge in the robotics domain (\textit{M} = 3.6, \textit{SD} = 1.7) but a lower experience with robots (\textit{M} = 2.6, \textit{SD} = 1.6), as indicated on a 7-point Likert scale from 1 = 'no knowledge / experience' to 7 = 'very knowledgeable / experienced'. Neither knowledge about nor experience with robots influenced any of measures conducted in our study (i.e., all correlations with the dependent variables were insignificant).

\section{Results}
To test our hypotheses, we ran a series of two-way ANOVAs with robot gender (male vs. female) and task type (analytical vs. social) as independent variables. Normality checks and Levene’s test were carried out and the assumptions were met.

\subsection{Trust}
\begin{figure}[!htbp]
      \centering
     \includegraphics[width=0.45\textwidth]{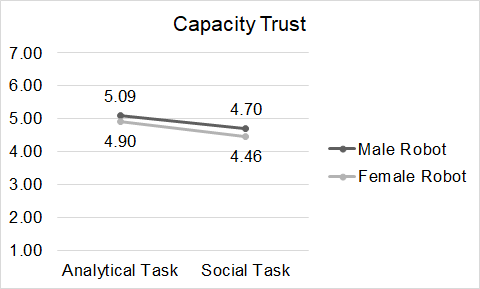}
      \caption{Robot gender vs. task type on capacity trust}
      \label{fig:CapacityTrust}
 \end{figure}

We observed a significant main effect for task type (\textit{F}(3,1) = 4.79, \textit{p} = .031, \textit{d} = .47) on capacity trust, but not for robot gender (\textit{F}(3,1) = 1.27, \textit{p} = .264, \textit{d} = .25) nor for their interaction effect (\textit{F}(3,1) 0.02, \textit{p} = .885, \textit{d} = .05). However, we observed no significant main effect for robot gender (\textit{F}(3,1) = 2.05, \textit{p} = .156, \textit{d} = .31) or task type (F(3,1) = 1.26, p = .264, \textit{d} = .25) on moral trust nor for their interaction effect (\textit{F}(3,1) 0.10, \textit{p} = .748, \textit{d} = .06). These results suggest that only people’s capacity trust in a robot is affected and exclusively by the type of task it performed. Specifically, participants have higher trust in a robot’s capacity when it performed an analytical task compared to when it performed a social task (see Figure \ref{fig:CapacityTrust}).

\subsection{Social Perception}
\begin{figure}[!htbp]
      \centering
      \includegraphics[width=0.45\textwidth]{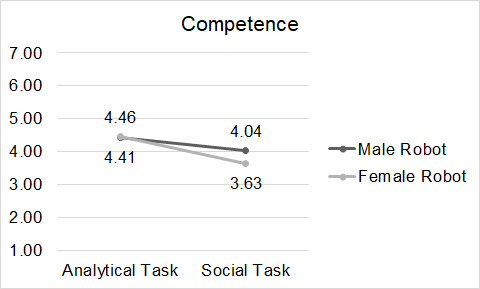}
      \caption{Robot gender vs. task type on competence}
      \label{fig:Competence}
 \end{figure}

We observed no significant main effect for robot gender (\textit{F}(3,1) = 1.26, \textit{p} = .265, \textit{d} = .25) or for task type (\textit{F}(3,1) = 0.19, \textit{p} = .666, \textit{d} = .09) on warmth nor for their interaction effect (\textit{F}(3,1) < 0.01, \textit{p} = .990, \textit{d} = .05). However, we observed a significant main effect for task type (\textit{F}(3,1) = 7.11, \textit{p} = .009, \textit{d} = .58) on competence, but not for robot gender (\textit{F}(3,1) = 0.62, \textit{p} = .434, \textit{d} = .17) nor for their interaction effect (\textit{F}(3,1) = 1.04, \textit{p} = .311, \textit{d} = .22). These results suggest that people’s social perception of a robot is particularly affected by the task it performs. Specifically, people perceive a robot as more competent when it performs an analytical task compared to a social task independent of the robot’s gender (see Figure \ref{fig:Competence}).

\subsection{Humanness}
\begin{figure}[!htbp]
      \centering
      \includegraphics[width=0.45\textwidth]{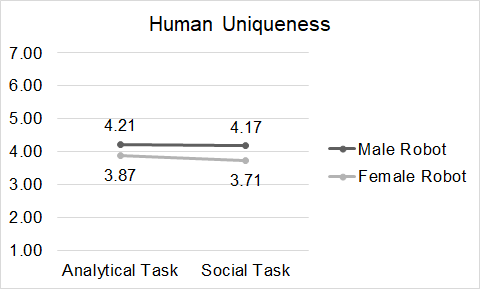}
      \caption{Robot gender vs. task type on human uniqueness}
      \label{fig:HumanUniqueness}
 \end{figure}

\begin{figure}[!htbp]
      \centering
      \includegraphics[width=0.45\textwidth]{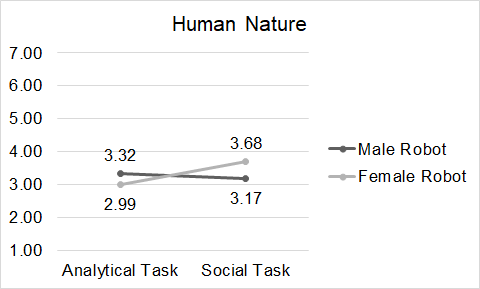}
      \caption{Robot gender vs. task type on human nature}
      \label{fig:HumanNature}
 \end{figure}

We observed a marginally significant main effect for robot gender (\textit{F}(3,1) = 2.77, \textit{p} = .100, \textit{d} = .35) on human uniqueness, but not for task type (\textit{F}(3,1) = 0.17, \textit{p} = .683, \textit{d} = .09) nor for their interaction effect (\textit{F}(3,1) = 0.06, \textit{p} = .812, \textit{d} = .06). Moreover, we observed no significant main effect for robot gender (\textit{F}(3,1) = 0.18, \textit{p} = .671, \textit{d} = .09) nor for task type (\textit{F}(3,1) = 1.51, \textit{p} = .223, \textit{d} = .28) on human nature while their interaction effect approached significance (\textit{F}(3,1) = 3.80, \textit{p} = .055, \textit{d} = .44). These results suggest that people’s humanness perception of a robot is independent of the robot’s gender and the type of task it performed, while the data indicated a trend where: (1) participants perceptions of a robot’s human uniqueness might be affected by robot gender; and (2) participants perceptions of a robot’s human nature might be an effect of the combination of robot gender and task type. Specifically, participants seem more inclined to dehumanize female robots to animals lacking higher-level mental processes (i.e., lacking human uniqueness) compared to male robots independent of the task they performed (see Figure \ref{fig:HumanUniqueness}). Moreover, participants seem to dehumanize robots to emotionless objects (i.e., lacking human nature) exclusively when a female robot performs analytical tasks or when a male robot performs social tasks (see Figure \ref{fig:HumanNature}).

\section{General Discussion}
Our study expands current knowledge on appearance-task fit and gender inferences by investigating the effects of robot genderedness (male vs. female) and assigned task (analytical vs. social) on social perception, trust, and humanness. 

Our data shows that people's evaluation of their trust in a robot only focuses on its capacity, but not on its morality, independent of its gender. These results show that trust evaluations of a robot cannot be linked to a robot's gender as we hypothesized (H2). Instead, our results indicate that trust in robots is more strongly associated with the performed task. Additionally, people perceive a robot to be more competent when it performs an analytical task compared to a robot performing a social task, independent of its gender. These results also contradict our hypothesis expecting an effect for robot gender on people’s social perception of a robot (H1). It seems that, when associating gendered robots with specific tasks, the observed effects of gender stereotyping in both the psychology \cite{bem1974measurement} and HRI \cite{eyssel2012s} research seem to steer away from the genderedness of the embodiment towards the (perhaps also perceived gender-stereotypical) performed tasks --at least in terms of our social perception of and trust in such robots. An earlier study examining the relationship among occupational gender-roles, user trust and gendered robots also found no significant difference in the perception of trust in the robot's competency when considering the gender of the robot \cite{bryant2020should}. Similar findings have been reported in our HRI studies on gender-task fit \cite{kuchenbrandt2014keep}, reporting that people are less willing to accept help from a robot with a typically female task (i.e., a social task). 

These combined results on the dominant effects for task type, rather than for robot gender, are not completely surprising though. Prior research indeed shows that people in general hold more utilitarian perceptions on robots \cite{enz2011social, graaf2016anticipating, goetz2003matching}. However, we also need to point to a potential limitation of our stimuli. Although the male robot was significantly perceived as more male than the female robot, its rating was leaning towards the female end of the gender scale. A similar observation was made for the social task, which was significantly perceived as less analytical than the analytical task yet its rating was leaning towards the analytical end of the scale. Future research should therefore not only explore different task descriptions but should also further investigate gendered appearances of robots or include of a gender-neutral robot as well as study subsequent (interaction) effects on people's social perception of and trust in such robots. Moreover, previous research in psychology \cite{buchan2008trust} as well as HRI \cite{siegel2009persuasive} shows interaction effects between the gender of the participant and that of the social other in terms of trust. Additional interaction effects between participant and robot gender have been found by \cite{otterbacher2017s} who reported a higher chance of an uncanny reaction to other-gender robots when that robot meets the gender expectations of warm females and competence males. Therefore, exploring interaction effects between the gender of the participant and the gender of robot in the context of gender-task fit might be promising as well.

Moreover, based on the literature in psychology, we expected that people’s perceptions of a robot’s humanness is a function of both robot gender and performed task (H3). This hypothesis was not supported with our data. However, we would like to discuss the observed trend in our data indicating a potential interaction effect between robot gender and performed task on a robot's perceived humanness. Following this trend, it seems that people dehumanize female robots (regardless of task performed) to animals lacking higher-level mental processes. Sexist responses to female robots have been reported in HRI research more generally \cite{strait2017public,veletsianos2008sex}. Additionally, people seem to dehumanize robots to emotionless objects only when gendered robots perform tasks contradicting the stereotypes of their gender. Research in social psychology has shown that women are dehumanized to both animals and objects \cite{rudman2012animals}, which is a precursor to aggressing against them \cite{haslam2006dehumanization}. Inserting the concept of gender into current debates regarding robot abuse (e.g., that mindless robots get bullied \cite{keijsers2018mindless}) might offer alternative perspectives on these issues. 

The field of social robotics aims to build robots that can engage in social interaction scenarios with humans in a natural, familiar, efficient, and above all intuitive manner \cite{de2015makes}. The easiest way to deal with expectations of gendered robots and subsequent stereotypical impressions is to enhance people’s social acceptance of gendered robots by tailoring their gender appearance to their intended task or application domain. Alternatively, robots might offer a unique potential to illuminate implicit bias in social cognition and challenging persisting gender-task stereotypes in society.


\bibliographystyle{ACM-Reference-Format}
\balance
\bibliography{references}










\end{document}